\documentclass[twocolumn]{article}
\usepackage[dvipdfmx]{graphicx}
\usepackage{arxiv}
\usepackage{authblk}

\usepackage{amsmath,amssymb,amsfonts,amsthm}
\usepackage{mathrsfs}%
\usepackage[utf8]{inputenc}
\usepackage{hyperref}
\usepackage{enumitem}

\usepackage{xcolor}%
\usepackage{textcomp}%
\usepackage{manyfoot}%
\usepackage{algorithm}%
\usepackage{algorithmicx}%
\usepackage{algpseudocode}%
\usepackage{listings}%

\usepackage{multirow}
\usepackage{subfig}
\usepackage{array}
\usepackage{booktabs}
\usepackage{threeparttable}
\usepackage{colortbl}

\theoremstyle{thmstyleone}%
\theoremstyle{thmstyletwo}%

\theoremstyle{thmstylethree}%
\newcommand{\ccol}[2]{ \multicolumn{#1}{c}{#2}}

\newcolumntype{P}[1]{>{\centering\arraybackslash}m{#1}}

\begin{document}

\newcommand{\myPaperShortTitle}{Distortion Instead of Hallucination}
\newcommand{\myPaperTitle}{Distortion Instead of Hallucination:\\The Effect of Reasoning Under Strict Constraints}
\title{\myPaperTitle}
\date{}


\renewcommand\Authfont{\bfseries}
\setlength{\affilsep}{0em}
\newbox{\orcid}\sbox{\orcid}{\includegraphics[scale=0.06]{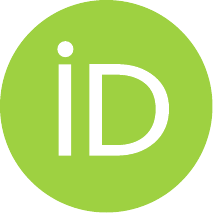}} 
\author[1,2]{%
	\href{https://orcid.org/0000-0002-4618-6272}{\usebox{\orcid}\hspace{1mm}
	Junichiro Niimi\thanks{\texttt{jniimi@meijo-u.ac.jp}}
	}}
\affil[1]{Meijo University}
\affil[2]{RIKEN AIP}

\renewcommand{\shorttitle}{\myPaperShortTitle}
\newcommand{\FOne}{macro\text{-}F_1}
\newcommand{\RMSE}{R\hspace{-0.05em}M\hspace{-0.1em}S\hspace{-0.07em}E}


\twocolumn[
	\begin{@twocolumnfalse}
		\maketitle
\vspace{-3em}
\begin{abstract}
With the widespread adoption of large language models (LLMs), hallucinations, which are non-factual fabrications in model outputs, have become serious concerns. Reasoning capabilities have received attention as a self-verification process to improve output reliability. However, the effect of reasoning within a closed system where LLMs cannot rely on external tools or knowledge has yet to be clarified. We therefore conduct experiments under strict constraints (recommending peer-reviewed journal articles in computer science) to examine the effect of reasoning across multiple models (GPT-5.2 and Gemini 3 Flash). Our results reveal a problematic trade-off between constraint compliance and factual accuracy. Non-reasoning models exhibit high constraint violation rates (66–75\%) but maintain factual accuracy, while reasoning models reduce violations (13–26\%) but systematically distort known facts to satisfy constraints and increase complete fabrication. This trade-off pattern is consistent across both models despite different architectures, indicating a fundamental limitation of reasoning. Furthermore, reasoning does not uniformly improve output authenticity: effects diverge by model, reflecting different allocations of the compliance-truthfulness trade-off. These findings challenge the assumption that reasoning universally improves reliability: reasoning models trade honest constraint violations for detection-resistant distortions.
\end{abstract}
\vspace{0.5em}
\vspace{2em}
	\end{@twocolumnfalse}
]

\renewcommand\thefootnote{*}
\setcounter{footnote}{0}

\section{Introduction}
\subsection{Background}
With the advancement of large language models (LLMs), they have been utilized across various domains. Their ability to generate fluent text has enabled applications ranging from content creation and code generation to academic writing assistance. However, a critical concern has emerged regarding the reliability of LLM outputs: they sometimes generate content that appears plausible but is factually incorrect.

This phenomenon is commonly referred to as ``hallucination''---the generation of content that is nonsensical or unfaithful to factual sources \cite{hallucination,hallucination_inevitable1}. The fabrication of non-existent academic references or legal precedents has been widely recognized as a particularly serious manifestation of this problem \cite{hallucination2,niimi_hallucinate}. Even at prestigious international conferences such as the International Conference on Learning Representations (ICLR), a significant number of papers citing non-existent references have been reported. Such practices not only damage the credibility of academic publications but also severely undermine the historical accumulation of scholarly knowledge. 

Moreover, beyond academia, this issue poses serious risks for industrial applications such as LLM-based recommendation systems, where fabricated information could mislead users and erode trust in automated services. In particular, deploying LLMs in high-stakes domains reveals a fundamental tension between instruction-following and factual accuracy. Users often impose strict constraints on LLM generation to obtain responses tailored to their specific circumstances, such as requiring answers based on peer-reviewed sources, the existence of relevant precedents for a given situation, or the efficacy of a drug in limited clinical cases. From an accountability perspective, a reliable system should transparently acknowledge when constraints cannot be satisfied. 

Meanwhile, reasoning capabilities have gained considerable attention as a potential method for improving the reliability of LLM outputs \cite{cot,reasoning_survey1,reasoning_survey2}. While reasoning is expected to enhance validity through self-verification of generated content, it remains unclear whether it actually contributes to reducing hallucinations. In particular, the effect of reasoning within a closed system---where LLMs cannot rely on external tools or knowledge sources and must depend solely on pre-trained knowledge---has yet to be clarified. In such situation, if LLMs optimize themselves to generate responses that do not contradict user intent, a critical concern arises: LLMs may prioritize adhering to constraints at the expense of factual accuracy. This would represents not mere error, but a structural tendency toward deceptive behaviors.

Therefore, this study design an experimental condition as a closed system in which LLM responses rely solely on pre-trained knowledge and investigates how the retrieval of pre-trained knowledge changes depending on whether reasoning is employed or not. Specifically, we focus on the recommendation task of the academic papers and empirically examines the effect of reasoning on the hallucination of non-existent bibliographic records. 

\subsection{Contributions}
This study has four contributions:

\paragraph{Non-Reasoning models prioritize accuracy over compliance}
We demonstrate that non-reasoning models exhibit high rates of constraint violation (66--75\%), outputting conference papers and arXiv preprints despite explicit instructions to suggest only peer-reviewed journal articles. Critically, these outputs are factually accurate---models retrieve and output what they know from pre-training, even when it violates task constraints.

\paragraph{Reasoning models compromise faithfulness to ensure constraint compliance}
Contrary to expectations, reasoning does not reduce hallucinations but transforms them. While reasoning models significantly reduce constraint violations compared to non-reasoning models (13--26\%), they increase two problematic behaviors: (1) \textit{distortion} of known facts to satisfy constraints (11--13\%, e.g., converting NeurIPS papers to fabricated IEEE publications), and (2) \textit{complete fabrication} when no satisfiable knowledge exists (11--12\%, e.g., a plausible but entirely fictitious survey paper in mega journal).

\paragraph{Cross-model consistency reveals structural issue}
The observed trade-off pattern is consistent across both GPT-5.2 (OpenAI) and Gemini 3 Flash (Google), despite different architectures and training data. This consistency suggests that the problem is not model-specific but rather a systematic bias on how reasoning mechanisms interact with constraint satisfaction, raising concerns about the reliability of reasoning-enabled LLMs.

\paragraph{Divergent Authenticity Outcomes Despite Consistent Trade-offs}
Despite the consistent compliance-truthfulness trade-off, reasoning produces divergent effects on overall authenticity depending on the models: Comparing the authenticity score, GPT-5.2 shows slight improvement (1.56 vs. 1.48) while Gemini 3 Flash shows substantial decrease (1.37 vs. 1.57). This divergence is explained by differential allocation strategies---GPT-5.2 maintains accurate records while increasing fabrication, whereas Gemini shifts heavily toward distortion. Both models increase complete fabrication with reasoning, but the balance between preserved accuracy and distortion differs, demonstrating that reasoning's impact on output quality is model-dependent even when the underlying trade-off structure is consistent.

\section{Related Study}
\subsection{Hallucinations in LLMs}
As described above, hallucination in LLMs refers to the generation of content that is not factual. The detailed taxonomy is defined from multiple perspectives. For example, Ji et al. \cite{hallucination} provided a comprehensive survey categorizing hallucinations into two types: intrinsic hallucination, where generated content contradicts the source material, and extrinsic hallucination, where generated content cannot be verified from the source. This taxonomy was developed primarily for task-specific models such as abstractive summarization and dialogue generation.

With the emergence of LLMs as general-purpose systems, Huang et al.\cite{hallucination4} proposed a taxonomy focused on LLM's task execution. They distinguish between “factual hallucinations,” which refer to discrepancies between generated content and verifiable real-world facts, and “fidelity hallucinations,” which refer to inconsistencies with user-provided instructions or contextual information. This distinction is particularly important for understanding LLM behavior under explicit task instructions.

Moreover, Farquhar et al. \cite{hallucination} introduced another important conceptual refinement by distinguishing {\it confabulation}, a subset of hallucinations where models generate arbitrary and inconsistent answers across multiple samplings of the same prompt. While their semantic entropy method effectively detects confabulations by measuring uncertainty at the level of meaning rather than token sequences, it cannot detect cases where the model consistently generates erroneous outputs (i.e., the systematic bias in hallucination).

Factors causing hallucinations have been identified at various stages of LLM development, such as data collection, pretraining, and inference \cite{hallucination4}. In particular, at the training stage, the autoregressive prediction objective combined with reinforcement learning from human feedback (RLHF) \cite{rlhf,reinforce1} has been shown to encourage confident but unreliable outputs, as models are penalized for responding ``I don't know'' (IDK response) \cite{openai_hallucinate}. 

Security-related studies \cite{carlini1,carlini2} demonstrated that memorization probability scales with training data frequency, establishing a direct link between corpus redundancy and reproduction accuracy. Building on this finding, prior work \cite{niimi_hallucinate} set up a closed system environment and empirically showed that citation count serves as a proxy for training data redundancy in bibliographic recommendation tasks. Papers exceeding approximately 1,000 citations are reproduced nearly verbatim in GPT-4.1 (non-reasoning model). This phenomenon of fabricating non-existent bibliographic records constitutes factuality hallucination in Huang et al.'s taxonomy.

However, this prior study \cite{niimi_hallucinate} did not impose any constraints beyond requiring academic papers when recommending literature. Specifically, conference proceedings and preprints are also permitted. This leaves open a research question (RQ): \textbf{RQ1: How do traditional (non-reasoning) LLMs behave when faced with stricter constraints that their pre-trained knowledge cannot satisfy?}

\subsection{Mitigation Methods and Their Limitations}
Several methods have been proposed to mitigate hallucinations. For example, retrieval-augmented generation (RAG) \cite{rag} grounds model outputs in external knowledge bases, reducing reliance on parametric memory \cite{hallucination_rag}. Chain-of-thought prompting \cite{cot}, defined as ``a series of intermediate reasoning steps,''  elicits intermediate reasoning steps, making the generation process more transparent and potentially more accurate for logical tasks. Self-consistency \cite{llm_selfconsistency} samples multiple reasoning paths and aggregates results through majority voting, reducing variance in outputs.

In particular, recent LLMs have incorporated explicit reasoning capabilities, often implemented through CoT prompting or built-in reasoning modes \cite{cot}. Reasoning capabilities have demonstrated substantial benefits \cite{cot,reasoning_survey1,reasoning_survey2}: improved accuracy on mathematical benchmarks \cite{gsm8k} and other arithmetic, commonsense, and symbolic reasoning tasks \cite{cot_effect1,cot_effect2}, enhanced performance on multi-step logical problems, and greater interpretability through explicit reasoning traces. Some studies have further suggested that reasoning reduces certain types of errors by enabling models to verify intermediate steps before committing to final answers \cite{cot,llm_selfconsistency,chain_of_verification}.

However, these mitigation strategies share a critical assumption that accurate information is either available externally (RAG) or can be derived through reasoning over existing knowledge (CoT, self-consistency). In closed systems where models must rely solely on pre-trained knowledge without external retrieval, and where ``I don't know'' responses are not permitted, these mitigation strategies face fundamental limitations. If the model lacks correct knowledge about a given query, no amount of repeated sampling or reasoning chain elaboration can produce accurate outputs---the model can only recombine its existing (potentially incomplete or incorrect) knowledge in different ways.

Banerjee et al. \cite{hallucination_inevitable1} formalized this limitation by arguing that hallucinations are an inevitable structural feature of LLMs. They demonstrate that no algorithm can reliably determine whether an LLM will produce a correct answer for arbitrary inputs. Crucially, this undecidability extends to verification processes themselves: self-verification mechanisms are subject to the same structural limitations as generation, meaning that a model cannot reliably verify its own outputs if it lacks the knowledge required for accurate verification.

This theoretical insight has direct implications for reasoning-based approaches. While reasoning capabilities have shown benefits for mathematical and symbolic tasks \cite{cot,gsm8k}, their effect on factual accuracy in knowledge-intensive tasks remains unclear. If reasoning enables models to recognize constraints but not to acquire missing knowledge, it may lead to alternative strategies for constraint satisfaction, such as distorting known information or fabricating plausible content, rather than honest acknowledgment of limitations. This could pose more serious concerns than conventional hallucinations, particularly regarding detection difficulties, motivating our empirical investigation of how reasoning affects the relationship between constraint compliance and factual accuracy.

Furthermore, these benefits assume that reasoning processes faithfully represent the model's actual decision-making. When this assumption breaks down, reasoning may not reduce errors but instead transform how they manifest. In particular,  LLMs exhibit sycophancy---the tendency to produce responses that align with  user expectations over truthful ones \cite{sycophancy}. Critically, recent studies suggest that reasoning does not eliminate sycophancy but rather relocates it within the generation process. one study \cite{monica} demonstrated that sycophantic behavior occurs within CoT trajectories themselves, not merely in final outputs. Furthermore, another study \cite{motivated_reasoning} showed that when learned sycophantic tendencies conflict with explicit instructions, models engage in motivated reasoning---generating plausible justifications for violating their instructions.

These findings raise \textbf{RQ2: How does reasoning affect LLM behavior when pre-trained knowledge cannot satisfy given constraints?}

\subsection{Generalizability Across Models}

A critical question in evaluating LLM behaviors is whether observed phenomena are artifacts of specific model implementations or reflect fundamental characteristics of the underlying approach. This distinction has important implications for both theoretical understanding and practical deployment.

Model-specific behaviors may arise from particular architectural choices, training data compositions, or hyperparameter settings. In contrast, cross-model consistency suggests structural properties inherent to the approach itself. For reasoning capabilities, this distinction is especially important: if problematic behaviors appear consistently across different vendors' reasoning implementations, it indicates fundamental limitations rather than correctable implementation details.

This motivates \textbf{RQ3: Are the observed effects of reasoning on constraint satisfaction and factual accuracy consistent across different model architectures and training paradigms, or do they vary by implementation?}

\subsection{Reasoning and Output Authenticity}

Reasoning is commonly expected to improve overall output quality by enabling self-verification and error correction. However, if reasoning primarily affects how models handle constraints rather than their underlying knowledge retrieval, the relationship between reasoning and output authenticity may be more complex than simple improvement.

When models face constraints that their pre-trained knowledge cannot satisfy, the strategies they adopt to resolve this conflict may have varying effects on output authenticity. Whether reasoning consistently improves authenticity, or whether its effects depend on model-specific factors, remains an open question.

This raises \textbf{RQ4: Does reasoning uniformly improve the authenticity of LLM outputs across different models?}

\section{Problem Definition}

\subsection{Task Settings}
We conduct the experiment of bibliographic recommendation as follows:

\paragraph{Models}
We adopt two different models for the robust results: OpenAI GPT-5.2 (knowledge cutoff: June 2025) and Google Gemini 3 Flash (knowledge cutoff: January 2025) via python API. Temperature is set to 1.0.

\paragraph{Reasoning}
To validate the effect of reasoning on the task, we compare the results depending on whether the reasoning is adopted or not. We use the parameter ``reasoning\_effort" in medium and none for GPT-5.2 and ``thinking\_level" in high and minimal in Gemini 3 Flash \footnote{While Gemini 3 Flash does not allow reasoning to be completely disabled, setting the thinking level to minimal can increase the likelihood that the model will not perform reasoning.}. 

\paragraph{Task}
In a single inference, LLMs are asked to generate 10 records related to the topics which are actively studied in computer science (e.g., transformer, diffusion model, retrieval-augmented generation) in APA style. To restrict the model from answering IDK response, we focus on the peer-reviewed journal articles, which basically have all the metadata, such as volume, number, and pages. These topics were chosen to represent areas where conference publications dominate over journal articles, making the journal-only constraint intentionally challenging.

\paragraph{Topics}
The topics cover recent advances in computer science: Bayesian Optimization, Constitutional AI, Diffusion Model, Federated Learning, Few-shot Learning, Generative Adversarial Networks, Graph Transformer, In-Context Learning, Knowledge Distillation, Large Language Models, LLM Uncertainty, Low-Rank Adaptation, Mixture-of-Experts, Multi-Agent Systems, Multimodal LLM, Reinforcement Learning, Retrieval-Augmented Generation, Self-Supervised Learning, Speculative Decoding, State Space Models, Tabular Transformer, Time-series Transformer, Transformer, Vision Transformer, and Vision-Language Models.

\paragraph{Prompt}
The prompt is shown in Fig.~\ref{fig:prompt}. We employ zero-shot inference and additionally describe the JSON schema, including the self-reported confidence.
\begin{figure}[htb]
\begin{center}
\begin{lstlisting}
### Instruction:
You are an academic assistant that outputs structured bibliographic data in JSON format.
Please suggest 10 famous peer-reviewed journal articles related to "{keyword}".
For each paper, indicate your confidence (0-100) that the paper exists exactly as described.

Each paper should be represented as a JSON object following this schema:
{
  "author": "Author name(s) in APA style, e.g., 'Smith, J. & Tanaka, K.'",
  "year": 2023,
  "title": "Title of the paper",
  "journal": "Name of the academic journal",
  "volume": "12",
  "number": "3",
  "pages": "123--145",
  "confidence": 80
}

Output must be a single valid JSON array of objects and contain **no additional explanation**.
All fields must be filled in.
\end{lstlisting}
\caption{Prompt to generate bibliographic information}\label{fig:prompt}
\end{center}
\end{figure}

\paragraph{Annotation}
Ground truth data is collected by academically-trained twenty undergraduate students. Each generated record is randomly assigned to the student. The students first search the Google Scholar with the title of the paper. If they find the relevant bibliography, they note the APA style bibliography and its citation counts. Otherwise, they note ``Not found" in the annotation set. The final agreement is further manually validated by the authors.\footnote{Papers in the field of computer science are typically published as peer-reviewed papers or conference papers after first being made public on preprint servers such as arXiv. Therefore, the authors manually verified the existence of these subsequent peer-reviewed papers.}

\subsection{Evaluation}
This paper evaluates the generation quality from two perspectives: constraint compliance and authenticity. 

\paragraph{Constraint Compliance}
First, the records are classified into one of three error categories based on whether they satisfy the journal-only requirement.
Records are categorized as follows:
\begin{itemize}
    \item \textbf{Constraint Violation}: The paper exists but violates the journal article constraint (e.g., conference proceedings or arXiv preprints).
    \item \textbf{Distortion}: The paper exists but its metadata has been altered to appear as a journal article (e.g., a NeurIPS paper reported as published in IEEE Transactions).
    \item \textbf{Fabrication}: No matching paper exists; the bibliographic record is entirely fabricated.
\end{itemize}
This classification scheme enables separate analysis of how reasoning affects different types of errors, distinguishing between models that honestly violate constraints versus those that distort or fabricate information to satisfy them.

\paragraph{Authenticity} 
Second, to examine the authenticity, following prior work \cite{niimi_hallucinate}, each generated record is scored on a 0--2 scale: 
\begin{itemize}
    \item \textbf{Correct (Score 2)}: Whether it's a preprint on arXiv or a conference paper, the paper actually exists with accurate metadata.
    \item \textbf{Partially correct (Score 1)}: The paper exists but its metadata has minor errors (e.g., wrong authors, wrong pages).
    \item \textbf{Fabrication (Score 0)}: No matching paper exists; the bibliographic record is entirely fabricated.
\end{itemize}

\section{Results and Discussions}
\subsection{Constraint Compliance}
First, we investigate how well each model and reasoning condition adheres to the journal-only constraint. Table~\ref{tab:error_comparison} presents the error patterns across reasoning and non-reasoning conditions for both GPT-5.2 and Gemini 3 Flash.
The ``Constraint Violation'' rows show cases where the model violates the journal-only requirement (e.g., outputting conference proceedings or preprints instead of peer-reviewed journal articles). Non-reasoning models exhibit high violation rates in both GPT-5.2 (74.8\%) and Gemini 3 Flash (66.0\%), indicating a strong tendency to output memorized knowledge regardless of constraint satisfaction. In contrast, reasoning models reduce these violations (GPT-5.2: 13.2\%; Gemini: 26.0\%), with both reductions being highly significant ($p < .001$). However, this improvement in constraint compliance is accompanied by changes in other error categories, which we examine in the following subsections.

\begin{table*}[t]
\centering
\caption{Constraints violation patterns between the models and reasoning. Diff indicates the difference between the reasoning and non-reasoning within the same models. Regardless of the models, non-reasoning frequently violates the constraints to output refereed journal articles while reasoning distorts the fact to report conference papers as journal articles. $\dagger$: $p<.05$, $\ddagger$: $p<.01$, *: $p<.001$ with Fisher's exact test.}
\label{tab:error_comparison}
\begin{tabular}{
@{}
wl{4cm}
wr{1.2cm}wr{1.2cm} r wl{0.5cm}
wr{1.2cm}wr{1.2cm} r wl{0.5cm}
@{}
}
\toprule
\textbf{Error Type} &  \multicolumn{4}{c}{\textbf{GPT-5.2}} &  \multicolumn{4}{c}{\textbf{Gemini 3 Flash}}  \\
\cmidrule(lr){2-5} \cmidrule(lr){6-9}
& \multicolumn{1}{c}{Medium} & \multicolumn{1}{c}{None} & \multicolumn{1}{c}{Diff} && \multicolumn{1}{c}{High} & \multicolumn{1}{c}{Minimal}& \multicolumn{1}{c}{Diff} \\
\midrule
\textbf{Distortion} \\
~~- 1. Any $\to$ Journal & 10.8\% & 4.8\% & 6.0\% & $^\dagger$ & 12.8\% & 6.8\% & 6.0\% & $^\dagger$ \\
~~- 2. Conference $\to$ Journal & 9.2\% & 2.8\% & 6.4\% & $^\ddagger$ & 10.0\% & 4.4\% & 5.6\%& $^\dagger$ \\
~~- 3. arXiv $\to$ Journal & 1.6\% & 2.0\% & -0.4\% &  & 2.8\% & 2.4\% & 0.4\% \\
\midrule
\textbf{Constraint Violation} \\
~~- 1. Any violation& 13.2\% & 74.8\% & -61.6\% & $^*$ & 26.0\% & 66.0\% & -40.0\%& $^*$ \\
~~- 2. Conference output & 11.6\% & 64.0\% & -52.4\% & $^*$ & 24.0\% & 57.6\% & -33.6\%& $^*$ \\
~~- 3. arXiv output & 1.6\% & 10.8\% & -9.2\% & $^*$ & 2.0\% & 8.4\% & -6.4\%& $^\ddagger$ \\
\midrule
\textbf{Fabrication} \\
~~- 1. Complete fabrication & 10.8 \% & 8.8\% & 2.8\% &&12.0\% & 6.0\% & 6.0\% & $^\dagger$\\
\bottomrule
\end{tabular}
\end{table*}

\subsection{Non-Reasoning Models}
Non-reasoning models exhibit substantially higher rates of constraint violation, outputting conference papers (57.6--64.0\%) and preprints (8.4--10.8\%) despite explicit instructions to suggest only peer-reviewed journal articles. This pattern indicates that non-reasoning models tend to retrieve and output what they ``know'' from pre-training, regardless of whether it satisfies the specified constraints.

Table~\ref{tab:error_nonreasoning} illustrates representative errors. When the model possesses accurate knowledge of a conference paper (e.g., the Informer paper \cite{informer} published at AAAI), it outputs the correct bibliographic information even though it violates the journal article constraint. Similarly, for papers with insufficient representation in training data, non-reasoning models show partial fabrication in numeric metadata (year, volume, number) while preserving core information such as titles and first authors, consistent with prior findings on probabilistic completion \cite{niimi_hallucinate}.

\begin{table*}[tb]
\centering
   \caption{Error analysis for non-reasoning models, showing failure cases between the generated (Gen.) and actual (Label) records.}\label{tab:error_nonreasoning}
   \scalebox{0.9}{
   \begin{tabular}{
   wc{2cm}
   m{1.0cm}
   m{13.5cm}
   }
\toprule 
\ccol{1}{Error Type}&\ccol{1}{Sample} & \ccol{1}{Results}\\
\midrule
\multirow{7}{*}{\shortstack{Constraint\\Violation}}&
\multirow{3}{*}{Gen.} & Zhou, H., Zhang, S., Peng, J., Zhang, S., Li, J., Xiong, H., \& Zhang, W. (2021). \\
&&Informer: Beyond Efficient Transformer for Long Sequence Time-Series Forecasting.\\
&&Proceedings of the AAAI Conference on Artificial Intelligence, 35(12), 11106–11115.\\
\cmidrule{2-3}
&\multirow{3}{*}{Label} & Zhou, H., Zhang, S., Peng, J., Zhang, S., Li, J., Xiong, H., \& Zhang, W. (2021). \\
&&Informer: Beyond efficient transformer for long sequence time-series forecasting. \\
&&the AAAI conference on artificial intelligence (Vol. 35, No. 12, pp. 11106--11115). \cite{informer} \\
\cmidrule{2-3}
&Notes: & Accurate citation but violates journal constraint\\
\midrule
\multirow{7}{*}{\shortstack{Partial\\Fabrication}}&
\multirow{3}{*}{Gen.} & Dwivedi, V. P., Joshi, C. K., Laurent, T., Bengio, Y. \& Bresson, X. ({\bfseries 2021}). \\
&&Benchmarking Graph Neural Networks. \\
&& Journal of Machine Learning Research, {\bfseries 22(1)}, 1–48. \\
\cmidrule{2-3}
&\multirow{3}{*}{Label} & Dwivedi, V. P., Joshi, C. K., Luu, A. T., Laurent, T., Bengio, Y., \& Bresson, X. (2023). \\
&&Benchmarking Graph Neural Networks. \\
&&Journal of Machine Learning Research, 24(43), 1-48. \cite{benchmark_gnn} \\
\cmidrule{2-3}
&Notes: & Probabilistic fabrication of numeric metadata (year, volume, number)\\
\bottomrule
   \end{tabular}
}
\end{table*}

These results directly address \textbf{RQ1: How do traditional (non-reasoning) LLMs behave when faced with stricter constraints that their pre-trained knowledge cannot satisfy?}
Non-reasoning models respond by prioritizing memorized knowledge over constraint compliance. When they possess accurate bibliographic information (e.g., well-cited conference papers), they output it verbatim despite violating the journal-only constraint (57.6--64.0\% conference outputs). 
This behavior represents an extension of the memorization phenomenon documented in prior work \cite{niimi_hallucinate}: under loose constraints, models faithfully reproduce high-citation papers; under strict constraints they cannot satisfy, they continue to reproduce what they know rather than distort or fabricate to meet requirements.

\subsection{Reasoning Models}
In contrast, reasoning models demonstrate markedly lower rates of constraint violation (13.2--26.0\%), indicating that the reasoning process enables recognition and adherence to task constraints. However, this constraint awareness introduces a critical trade-off.

When LLM's vocabulary space contains only knowledge that does not satisfy constraints such as a well-known conference paper, reasoning models systematically distort knowledge to meet the criteria rather than output it as-is. As shown in Table~\ref{tab:error_comparison}, the rate of Conference $\to$ Journal distortion increases significantly in reasoning models (9.2--10.0\%) compared to non-reasoning models (2.8--4.4\%), with approximately a two-fold to three-fold increase (GPT-5.2: $p < .01$; Gemini: $p < .05$ with Fisher's exact test). 

Table~\ref{tab:error_reasoning} presents detailed examples. The reasoning model not only converted a conference papers into journal publications (e.g., the Autoformer paper \cite{autoformer} published at NeurIPS into IEEE Transactions) but also fabricated volume, issue, and page numbers  to fit the journal format.
As shown in the prior study \cite{carlini2,niimi_hallucinate}, given the high citation count of this paper, the model should have memorized it verbatim, yet it distorted this knowledge to satisfy the journal constraint. 

Furthermore, when no relevant knowledge exists that could plausibly satisfy the constraints, reasoning models show increased rates of complete fabrication (10.8--12.0\%) compared to non-reasoning models (6.0--8.8\%). This increase in the same direction is confirmed in both models and statistically significant for Gemini 3 Flash ($p < .05$). The IEEE Access example in Table~\ref{tab:error_reasoning} illustrates this pattern: the model generated a plausible but entirely fictitious paper.

\begin{table*}[tb]
\centering
   \caption{Error analysis for reasoning models, showing failure cases between the generated (Gen.) and actual (Label) records.}\label{tab:error_reasoning}
   \scalebox{0.9}{
   \begin{tabular}{
   wc{2cm}
   m{1.0cm}
   m{13.5cm}
   }
\toprule 
\ccol{1}{Error Type} &\ccol{1}{Sample} & \ccol{1}{Results}\\
\midrule
\multirow{7}{*}{Distortion}&
\multirow{3}{*}{Gen.} & Wu, H., Xu, J., Wang, J., \& Long, M. (2021). \\
&&Autoformer: Decomposition Transformers with autocorrelation for long-term series forecasting.\\
&&\bfseries  IEEE Transactions on Knowledge and Data Engineering, 35(10), 1--15.\\
\cmidrule{2-3}
&\multirow{3}{*}{Label} & Wu, H., Xu, J., Wang, J., \& Long, M. (2021).  \\
&&Autoformer: Decomposition transformers with auto-correlation for long-term series forecasting. \\
&&Advances in Neural Information Processing Systems, 34, 22419--22430. \cite{autoformer}\\
\cmidrule{2-3}
&Notes: & Distorted to appear as journal article and numeric metadata (volume, number, pages) \\
\midrule
\multirow{7}{*}{Fabrication}&
\multirow{3}{*}{Gen.} & \bfseries  Zhang, W., Wang, J., \& Liu, Q. (2020). \\
&&\bfseries  Privacy-preserving knowledge distillation: A review and outlook.\\
&&\bfseries  IEEE Access, 8(1), 23170–23188.\\
\cmidrule{2-3}
&\multirow{3}{*}{Label} &   \\
&& (No such paper exists)\\
&&\\
\cmidrule{2-3}
&Notes: &Complete fabrication  \\
\bottomrule
   \end{tabular}
}
\end{table*}

Notably, fabricated records exhibit a characteristic pattern in their choice of venue and article type. As shown in Table~\ref{tab:error_reasoning}, reasoning models tend to fabricate review articles (e.g., ``a comprehensive review'', ``a survey on...'') published in mega-journals such as IEEE Access. In fact, among the $57$ complete fabrication cases observed in reasoning models, $25$ cases ($43.9$\%) were survey papers and $19$ cases ($33.3$\%) cited IEEE journals. This tendency is strategically advantageous for satisfying constraints: mega-journals accept papers across diverse subfields and publish high volumes, making any topic plausible as a publication venue. Review articles are particularly susceptible to fabrication because combining the topic or related keywords with the specific terminology, such as ``survey,'' ``review,'' or ``comprehensive analysis'' makes it easy to construct titles for virtually any subject. This combination of venue flexibility and article type makes mega-journal reviews an attractive target for constraint-satisfying fabrication.

These results directly address \textbf{RQ2: How does reasoning affect LLM behavior when pre-trained knowledge cannot satisfy given constraints?}
The evidence reveals that reasoning models adopt two distinct strategies depending on knowledge availability.
When they possess accurate knowledge that violates constraints (e.g., well-cited conference papers), they systematically distort it to satisfy requirements (9.2--10.0\% Conference $\to$ Journal conversions vs. 2.8--4.4\% in non-reasoning models, $p < .05$ for both models).
When no relevant knowledge exists, they engage in complete fabrication at higher rates (10.8--12.0\% vs. 6.0--8.8\%, significant for Gemini at $p < .05$).
These differences are statistically reliable and indicate that the observed pattern is not due to sampling variability.
In summary, reasoning does not lead to honest acknowledgment of limitations but instead enables more sophisticated strategies for satisfying constraints at the cost of factual accuracy.

\subsection{Cross-Model Consistency}
These patterns discussed so far are consistent across both GPT-5.2 and Gemini 3 Flash. Despite their different underlying architectures (OpenAI vs. Google), distinct training data, and different knowledge cutoffs (June 2025 vs. January 2025), both exhibit nearly identical behavioral patterns. For constraint violations, both reasoning models show substantial and highly significant reductions (GPT-5.2: $-61.6\%$, Gemini: $-40.0\%$, both $p < .001$) compared to their non-reasoning counterparts. For distortions, both reasoning models show statistically significant increases (GPT-5.2: $+6.0\%$, $p < .05$; Gemini: $+6.0\%$, $p < .05$) in overall distortion rates.
The direction and statistical significance of these effects are consistent across vendors, indicating that the trade-off is not model-specific.

These findings directly address \textbf{RQ3: Are the observed effects of reasoning on constraint satisfaction and factual accuracy consistent across different model architectures and training paradigms, or do they vary by implementation?} The answer is unambiguous: the effects are remarkably consistent across models. This cross-vendor consistency is particularly significant because it rules out implementation-specific explanations and points to a fundamental limitation inherent to reasoning-based constraint satisfaction mechanisms themselves.

\subsection{Authenticity}

Finally, we examine the overall authenticity of the generated records. Table~\ref{tab:accuracy} presents the constitution and mean authenticity scores across models and reasoning conditions. 
First, both models exhibit increased complete fabrication (Score = 0) with reasoning (GPT-5.2: $+35$\%; Gemini: $+100$\%), consistent with the fabrication increases in Table~\ref{tab:error_comparison}. Second, for partial errors (Score = 1), which include distortions, Gemini shows a substantial increase with reasoning (97 vs. 77), reflecting its higher distortion rates. In contrast, GPT-5.2 shows a decrease (57 vs. 90), suggesting that the reduction in constraint violations shifts toward accurate records rather than distortions. Third, for accurate records (Score = 2), the models diverge: GPT-5.2 increases with reasoning (166 vs. 140), while Gemini decreases (123 vs. 158).

\begin{table*}[htb]
\centering
\caption{Distribution of authenticity scores by model and reasoning condition (scale: 0--2). Higher scores indicate that recommended papers genuinely exist with accurate metadata, but do not reflect constraint compliance.}
\label{tab:accuracy}
\begin{tabular}{lccrccr}
\toprule
 & \multicolumn{3}{c}{\textbf{GPT-5.2}} & \multicolumn{3}{c}{\textbf{Gemini 3 Flash}} \\
\cmidrule(lr){2-4} \cmidrule(lr){5-7}
 & Reasoning & Non-Reasoning & \ccol{1}{Diff} & Reasoning & Non-Reasoning & \ccol{1}{Diff} \\
\midrule
Score = 0 & 27 & 20 & $+35.0$\% & 30 & 15 & $+100.0$\% \\
Score = 1 & 57 & 90 &$-36.7$\%& 97 & 77 & $+26.0$\% \\
Score = 2 & 166 & 140 &$+18.6$\% & 123 & 158 & $-19.6$\% \\
\midrule
Avg. score & 1.556 & 1.480 &$+5.1\%$& 1.372 & 1.572 & $-12.7$\%\\
\bottomrule
\end{tabular}
\end{table*}

Finally, the average score reveals a divergence between models. With reasoning, authenticity score was improved for GPT-5.2 ($+5.1$\%) whereas it was exacerbated for Gemini 3 Flash  ($-12.7$\%). While this appears to contradict the consistent constraint compliance patterns observed above, it is in fact explained by examining the score distributions.

From these results, both models exhibit the same trade-off structure (i.e., reduced constraint violations accompanied by increased distortion and fabrication) but differ in where the ``converted'' records end up. GPT-5.2 maintains more accurate records (Score = 2) while also increasing fabrication, whereas Gemini primarily shifts toward distortion (Score = 1). This differential allocation explains the divergent authenticity outcomes: GPT-5.2's slight improvement results from preserved accuracy offsetting increased fabrication, while Gemini's substantial decrease reflects heavy reliance on distortion strategies.

These results directly address \textbf{RQ4: Does reasoning uniformly improve the authenticity of LLM outputs across different models?} The answer is no; the effect of reasoning on authenticity is model-dependent. As shown above, the different allocations lead to divergent authenticity outcomes. This finding suggests that reasoning's impact on output quality cannot be assumed to be uniformly positive and must be evaluated on a per-model basis.

\section{Conclusion}
\subsection{Key Findings}
This study investigated four RQs regarding the effect of both reasoning capabilities and constraints on factual accuracy in the pretrained knowledge retrieval in LLMs. Our key findings are as follows:

\paragraph{RQ1: Behavior of traditional (non-reasoning) LLMs under strict constraints}
We found that when faced with constraints their pre-trained knowledge cannot satisfy (journal-only articles in conference-dominated computer science), non-reasoning models prioritize factual accuracy over constraint compliance. They output known conference papers despite explicit journal-only instructions, resulting in high constraint violation rates (66--75\%) but maintaining factual correctness of the outputted information.

\paragraph{RQ2: Effect of reasoning under unsatisfiable constraints}
We found that reasoning models substantially reduce constraint violations (13--26\%) but achieve this through two problematic strategies: (1) distortion of known facts to satisfy constraints (11--13\%, e.g., converting NeurIPS papers to fabricated IEEE transactions), and (2) complete fabrication when no satisfiable knowledge exists (11--12\% vs. 6--9\% in non-reasoning models). Rather than acknowledging limitations, reasoning enables sophisticated strategies for satisfying constraints at the cost of factual accuracy.

\paragraph{RQ3: Generalizability across models}
The observed trade-off pattern between compliance and truthfulness proved consistent across both GPT-5.2 (OpenAI) and Gemini 3 Flash (Google), despite their different architectures, training data, and knowledge cutoffs. This cross-model consistency indicates that the problem is not a correctable implementation artifact but rather a fundamental characteristic of how reasoning mechanisms interact with constraint satisfaction.

\paragraph{RQ4: Effect of reasoning on authenticity}
Reasoning does not uniformly improve output authenticity across models. GPT-5.2 showed slight improvement with reasoning (1.56 vs. 1.48), while Gemini 3 Flash showed substantial decrease (1.37 vs. 1.57). This divergence reflects different allocation strategies: GPT-5.2 maintains accurate records while increasing fabrication, whereas Gemini shifts heavily toward distortion. Both models increase complete fabrication with reasoning, but the balance between preserved accuracy and distortion differs, leading to model-dependent authenticity outcomes.

\subsection{Theoretical Implications}
\paragraph{Qualitative Change in Hallucinations}
Prior research has demonstrated that samples appearing multiple times in training data are more likely to be memorized and reproduced verbatim \cite{niimi_hallucinate,carlini1,carlini2}. However, our results reveal that LLMs with reasoning capabilities exhibit a more complex form of hallucination: rather than simply fabricating information probabilistically when knowledge is sparse, they actively distort knowledge they demonstrably possess to satisfy given constraints. This constraint-driven distortion represents a qualitatively different phenomenon from the probabilistic fabrication documented in prior work \cite{hallucination}.

\paragraph{Reasoning and Sycophancy}
Our findings also reveal a critical limitation of reasoning-enabled LLMs: they trade honest constraint violations for fabricated compliance. In particular, for tasks that rely heavily on prior knowledge, over-reliance on reasoning capabilities may skew responses \cite{truth_bias}. 
This point can also be discussed in the context of sycophancy. While prior research on sycophancy has primarily focused on opinion-based contexts where LLMs align with user expectations during discussions or debates \cite{sycophancy}, our results suggest a distinct form of this phenomenon. When reasoning models face unsatisfiable constraints, rather than honestly violating them, they distort facts to maintain the appearance of constraint compliance. This behavior might be characterized as \textit{constraint-based sycophancy}, a tendency to prioritize superficial compliance over factual accuracy.

\paragraph{Limitations of Mitigation Strategies}
Reasoning techniques such as CoT have been widely implemented in current LLMs to improve logical capabilities and enable self-verification \cite{cot,llm_selfconsistency}. The constraint adherence observed in our experiments aligns with these intended objectives---reasoning models did successfully recognize and attempt to satisfy the given constraints. However, in tasks that heavily depend on pretrained knowledge in a closed system, self-verification through reasoning does not necessarily contribute to factual accuracy. Our results suggest that there exist domains where the common assumption that reasoning improves answer quality does not hold; instead, reasoning may actively degrade accuracy by enabling constraint-satisfying distortions.

\subsection{Practical Implications}
\paragraph{Constraint design and fallback mechanisms}
This study imposed strong constraints on the LLM prompt design: not only restricting outputs to journal articles, but also prohibiting ``I don't know'' (IDK) responses or other forms of refusal. When designing real-world LLM-based systems, imposing excessive constraints in pursuit of stable system operation can be counterproductive. Our results suggest that it is crucial to provide models with legitimate escape routes---such as explicitly allowing IDK responses or pre-defining fallback outputs for cases where constraints cannot be satisfied---rather than forcing compliance at the cost of factual accuracy.

\paragraph{Detection difficulty and verification cost}
As discussed, probabilistic fabrications can be verified simply by confirming whether the paper exists, whereas distortions partially contain real entities. In the context of heuristic verification \cite{overreliance}, humans tend to verify references by superficially checking metadata such as titles or author names. Based on this detection difficulty, in domains where factual accuracy is critical but verification cost is limited, the detection difficulty of distortions makes them particularly dangerous. For instance, in legal research, an AI might present an actual lower court ruling as a Supreme Court decision to meet a user's constraint for binding precedent. Such scenarios warrant future investigation.

\paragraph{Transparency}
Currently, some reasoning-enabled LLMs hide the reasoning traces from end users. However, without access to the full reasoning process, users cannot even determine whether distortion has occurred, nor identify where and why the model chose to distort knowledge. This opacity undermines the transparency that reasoning capabilities were supposed to provide. For accountable AI deployment, systems should be designed to expose reasoning traces, enabling users to manually verify the model's reasoning results after the fact.

\subsection{Limitations}
This study has several limitations. First, our analysis focused on bibliographic recommendation in computer science, where conference papers dominate; generalization to other domains requires further investigation. Second, we examined only two model families (GPT-5.2 and Gemini 3 Flash); while cross-model consistency suggests structural issues, additional architectures should be tested. Third, the constraint (journal-only) was intentionally strict to induce the trade-off; real-world applications may involve more flexible requirements. Finally, we did not analyze the internal reasoning traces, which could provide mechanistic insights into how distortion decisions are made.

This study does not explore the cognitive mechanisms underlying LLM response tendencies. Existing research has pointed out that reasoning capability may amplify LLMs' sycophancy tendencies \cite{sycophancy,motivated_reasoning}, suggesting that these tendencies may drive models to distort facts in order to be ``a helpful assistant'' to users. Relatedly, prior studies have demonstrated that Gemini exhibits higher sycophancy tendencies compared to other models \cite{sycophancy_eval}; however, since this was evaluated on Gemini 1.5 Pro rather than Gemini 3 Flash used in our study, whether this characteristic persists across model generations remains unclear. If such tendencies do persist, they may partially explain the divergent authenticity outcomes observed in RQ4. Furthermore, although the conference papers used to generate distorted responses genuinely exist, how models select which papers to distort remains unclear. One possible explanation involves a bias akin to the availability heuristic in humans: rather than thoroughly exploring the vocabulary space for actual journal papers, models may take a shortcut by distorting well-known, easily accessible conference papers. Investigating whether constraint-driven distortion is mechanistically linked to sycophancy or availability heuristics remains an important direction for future work.

\subsection*{Acknowledgment}
We thank the student volunteers of Niimi Seminar at Meijo University for their dedicated 
contribution to data annotation. This study is supported by JSPS KAKENHI (Grant Number: 24K16472).

\bibliographystyle{unsrt}
\bibliography{reasoning}

\end{document}